\documentclass{hld2026} 

\usepackage[makeroom]{cancel}
\usepackage{algorithm,algpseudocode} 
\usepackage{wrapfig}
\usepackage{caption}
\usepackage{titlesec}
\usepackage{newtxtext,newtxmath}
\usepackage{float}
\usepackage{titletoc}
\usepackage{microtype}
\usepackage{comment}
\usepackage{booktabs}
\usepackage{array}
\usepackage{enumitem}
\usepackage{chngcntr}

\title[Dominant Fluctuations Lower Sharpness]{Mini-batch Noise Lowers Sharpness via Dominant-Subspace Fluctuations}

\hldauthor{%
 \Name{Junho So} \Email{jhso@ajou.ac.kr}\\
 \Name{Dongwook Shin} \Email{dws@ajou.ac.kr}\\
 \addr {Department of Mathematics, Ajou University, South Korea} 
}

\begin{document}

\vspace*{-5mm}
\maketitle
\vspace*{-8mm}

\begin{abstract}

During SGD training, the gradients often align strongly with the dominant subspace spanned by the top-\(k\) eigenvectors of the Hessian of the loss. While this seems to naturally imply that loss reduction mainly occurs within this space, prior work has shown that updates within this dominant subspace make no meaningful progress in reducing the loss.
In this work, we argue that the dominant subspace is better understood not as the main space for loss reduction, but as a key subspace for explaining the sharpness dynamics of mini-batch SGD. To explain the role of the dominant subspace in reducing top-\(k\) sharpness, we show how the averaged gradient over fluctuations in the dominant directions produces a sharpness correction term, and derive a sharpness correction term induced by mini-batch noise in the dominant directions. Experimental results show that adding the derived correction term to GD brings the sharpness evolution of GD closer to that of SGD.

\end{abstract}

\section{Introduction}
\label{sec:intro}


 Understanding the dynamics of training deep neural networks is one of the main topics of machine learning. During stochastic gradient descent (SGD) training, it has been observed that the Hessian of the loss often has a small number of large outlier eigenvalues~\citep{sagun2017eigenvalues,DBLP:journals/corr/SagunEGDB17,pmlr-v97-ghorbani19b,pmlr-v97-papyan19a,JMLR:v21:20-933}, and training gradients are known to align strongly with the \textit{dominant subspace}, which is spanned by the top-\(k\) Hessian eigenvectors~\citep{gurari2018gradientdescenthappenstiny,pmlr-v97-ghorbani19b}.
This suggests that the effective training dynamics of SGD may be low-dimensional, and it naturally leads to the expectation that loss reduction mainly occurs within this space.

However, prior work~\citep{song2025does} reports results that contradict this interpretation.
Specifically, when the SGD update is projected onto the dominant subspace, training achieves no meaningful loss reduction. In contrast, when the update is projected onto the \textit{bulk subspace}, which is the orthogonal complement of the dominant subspace, the loss decreases similarly to standard SGD. Does the dominant subspace therefore contribute nothing to training?

In this paper, we argue that the main role of the dominant subspace lies in reducing the top-\(k\) sharpness of the loss landscape. Specifically, we observe that the dominant-projected update significantly reduces sharpness even though it has little impact on loss reduction. Then, through experiments involving controlled perturbations along different subspaces (\textit{dominant} and \textit{random}), we show that only perturbations in the dominant directions reduce sharpness. To explain this effect, we first show that averaging the gradient over fluctuations in the dominant directions yields a deterministic sharpness correction term. We then derive the deterministic correction term induced by mini-batch noise in the dominant directions. Finally, we empirically show that adding this correction term to GD produces sharpness dynamics similar to SGD.

\vspace{-4mm}
\section{Setup and Motivation}
\label{sec:setup}
\vspace{-1mm}

\begin{figure}[!t]
  \centering
  \includegraphics[width=0.32\textwidth]{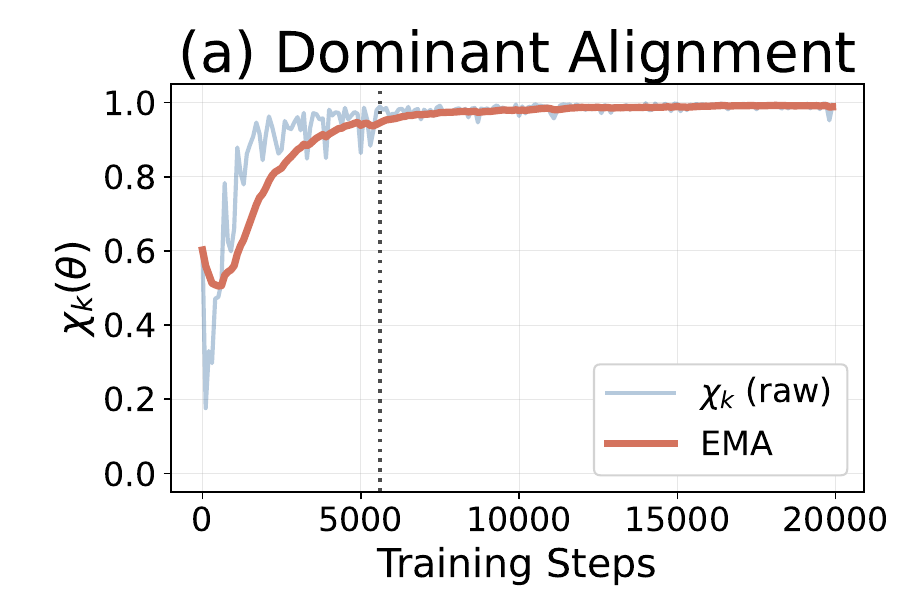}
  \includegraphics[width=0.32\textwidth]{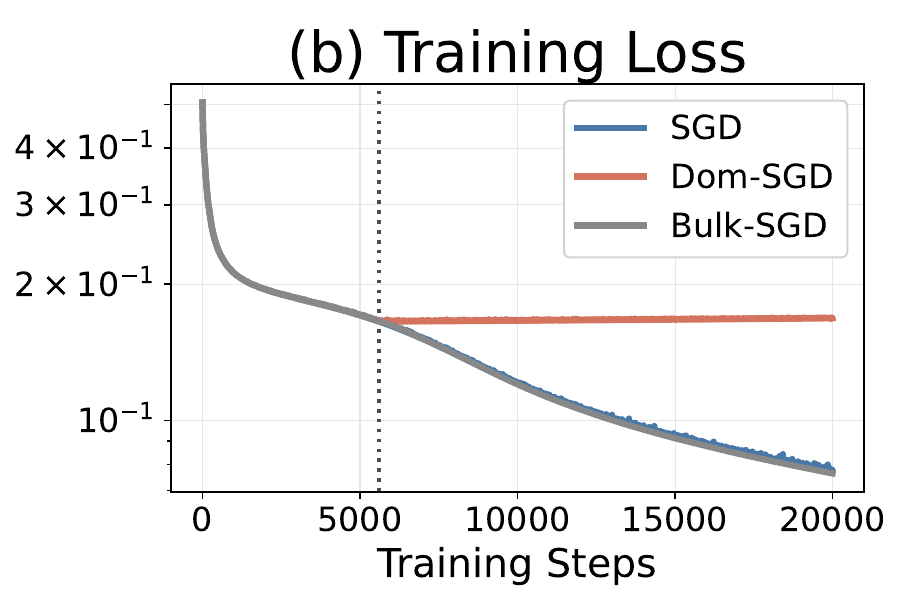}
  \includegraphics[width=0.32\textwidth]{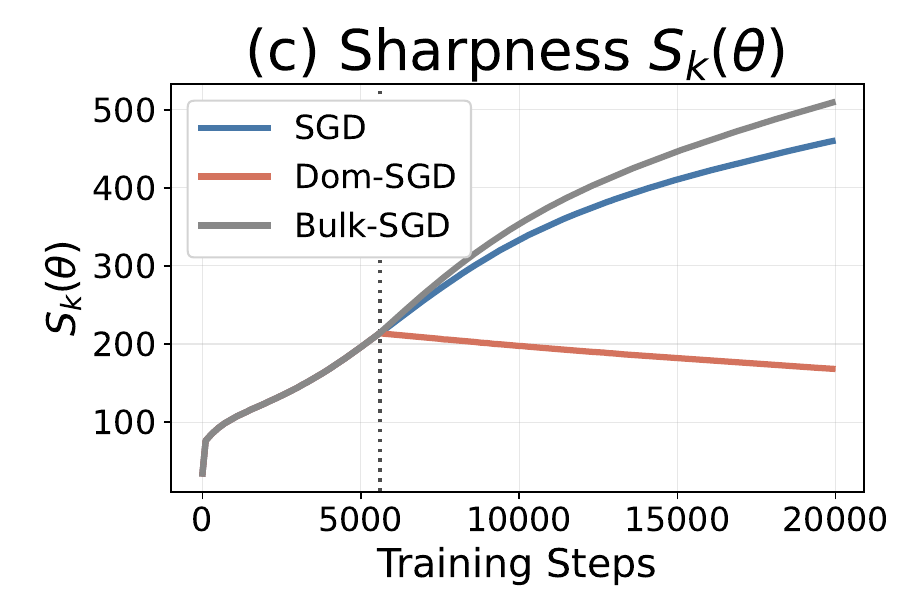}

    \caption{
\textbf{The dominant component does not reduce loss but reduces sharpness.}
(a) Training gradients align with the dominant subspace.
(b) Dom-SGD fails to reduce loss, while Bulk-SGD continues to train.
(c) Dom-SGD lowers \(S_k\), whereas Bulk-SGD maintains higher \(S_k\) than SGD.
}
  \label{fig:projection}
\end{figure}

\paragraph{Setup.}
Let \(L:\mathbb{R}^d\to\mathbb{R}\) be the loss function.
For a mini-batch \(B\) with mini-batch loss \(L_B\), define the mini-batch noise as
\(\xi_B(\theta):=\nabla L_B(\theta)-\nabla L(\theta)\), where
\(\mathbb{E}_B[\xi_B(\theta)\mid\theta]=0\).
We write mini-batch SGD as
\(\theta_{t+1}=\theta_t-\eta(\nabla L(\theta_t)+\xi_{B_t}(\theta_t))\),
and full-batch GD as
\(\theta_{t+1}=\theta_t-\eta\nabla L(\theta_t)\).
Let \((\lambda_i(\theta), e_i(\theta))\) be the eigenpairs of
\(H(\theta)=\nabla^2 L(\theta)\), ordered so that
\(\lambda_1(\theta)\ge\cdots\ge\lambda_d(\theta)\).
We define the \textit{dominant subspace} as the space spanned by the top-\(k\) eigenvectors, \(E_{\mathrm{dom}}(\theta)=\mathrm{span}\{e_1(\theta),\ldots,e_k(\theta)\}\), and define its orthogonal complement \(E_{\mathrm{bulk}}(\theta)=E_{\mathrm{dom}}(\theta)^\perp\) as the \textit{bulk subspace}. We also define the projections onto these spaces as the dominant projection \(P_{\mathrm{dom}}(\theta)\) and the bulk projection \(P_{\mathrm{bulk}}(\theta)=I-P_{\mathrm{dom}}(\theta)\). We define top-\(k\) sharpness as
\(S_k(\theta)=\sum_{j=1}^k \lambda_j(\theta)\). We restrict our focus to the stable learning-rate regime, where \(\lambda_1(\theta)<2/\eta\) is maintained during training.

 \vspace{-2mm} 
\subsection{Motivation and Main Observation} 
 \vspace{-1mm} 

Let us define the dominant alignment metric
\(\chi_k(\theta)=\|P_{\mathrm{dom}}(\theta)\nabla L(\theta)\|/\|\nabla L(\theta)\|\),
which quantifies the relative magnitude of the training-loss gradient lying in the dominant subspace along the SGD trajectory. As shown in Figure~\ref{fig:projection}(a), the SGD training gradient is strongly concentrated in the dominant subspace from early in training, and this tendency becomes stronger at later stages \citep{gurari2018gradientdescenthappenstiny,pmlr-v97-ghorbani19b}. This observation suggests that SGD dynamics are closely related to the dominant subspace.

\begin{wrapfigure}[8]{r}{0.24\textwidth}
   \vspace{-5mm}
  \centering
  \includegraphics[width=0.21\textwidth]{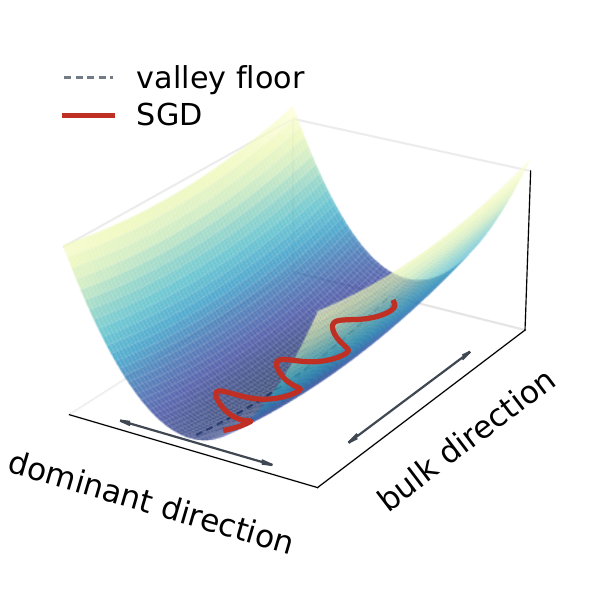}
  \vspace{-2mm}
  \captionsetup{font=scriptsize}
  \caption{River-valley intuition.}
  \label{fig:valley}
  \vspace{-5mm}
\end{wrapfigure}

However, as shown in Figure~\ref{fig:projection}(b), Dom-SGD, which projects the SGD update onto the dominant subspace, does not substantially reduce the training loss. Instead, Bulk-SGD, which keeps only the bulk component, reduces the loss as effectively as SGD~\citep{song2025does}. Prior work~\citep{song2025does,wen2025understanding} explains this phenomenon using the river-valley intuition. In this view, as shown in Figure~\ref{fig:valley}, the dominant directions correspond to the high-curvature valley walls, where updates do not reduce the loss, whereas the bulk directions follow the valley floor, where loss reduction mainly occurs.

Nevertheless, prior work has reported that models trained with Bulk-SGD alone fail to fully recover the final accuracy gains achieved by SGD \citep{zakarin2025acceleratingneuralnetworktraining}. This suggests that the dominant subspace, while not primarily responsible for loss reduction, may still play a meaningful role in training. Its contribution may instead lie in aspects of the optimization dynamics that are not directly captured by loss reduction alone.

In this work, we argue that the dominant subspace plays an important role in reducing the top-\(k\) sharpness of the loss landscape.
As shown in Figure~\ref{fig:projection}(c), Bulk-SGD follows almost the same loss curve as SGD, but its top-\(k\) sharpness becomes higher than that of SGD.
In contrast, Dom-SGD keeps the loss high while substantially reducing top-\(k\) sharpness, suggesting that the dominant subspace is connected to sharpness reduction rather than loss reduction. In the next section, we present experiments examining whether motion along the dominant directions is in fact the factor that reduces top-\(k\) sharpness.


\vspace{-3mm}
\section{Dominant-Subspace Fluctuations Reduce Sharpness}
\label{sec:main}

  \begin{figure}[!t]
    \centering
    \includegraphics[width=0.32\textwidth]{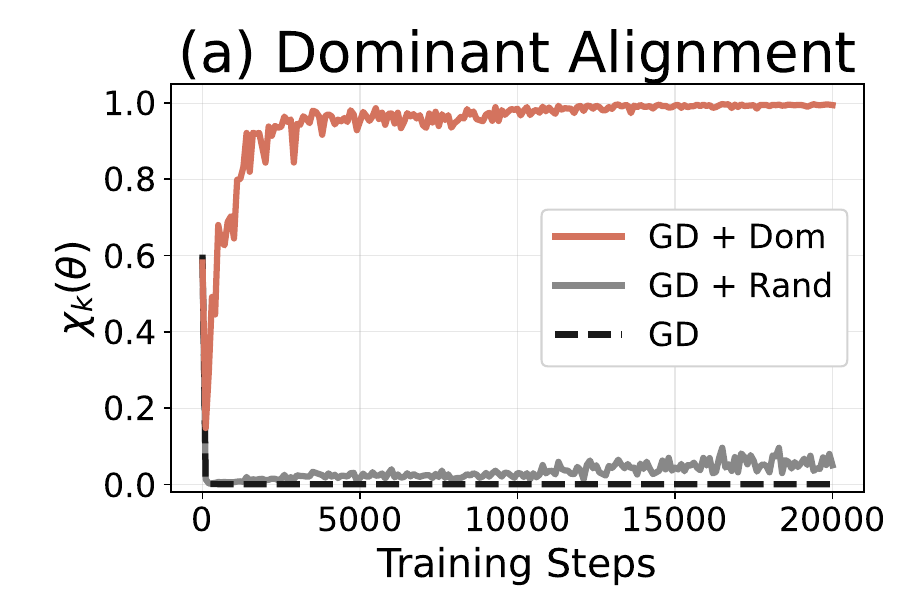}\hfill
    \includegraphics[width=0.32\textwidth]{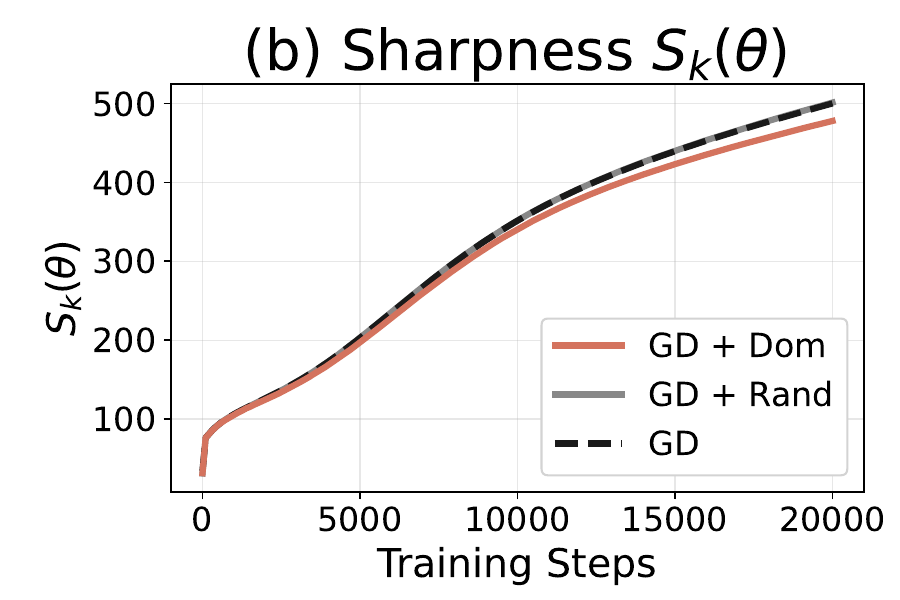}\hfill
    \includegraphics[width=0.32\textwidth]{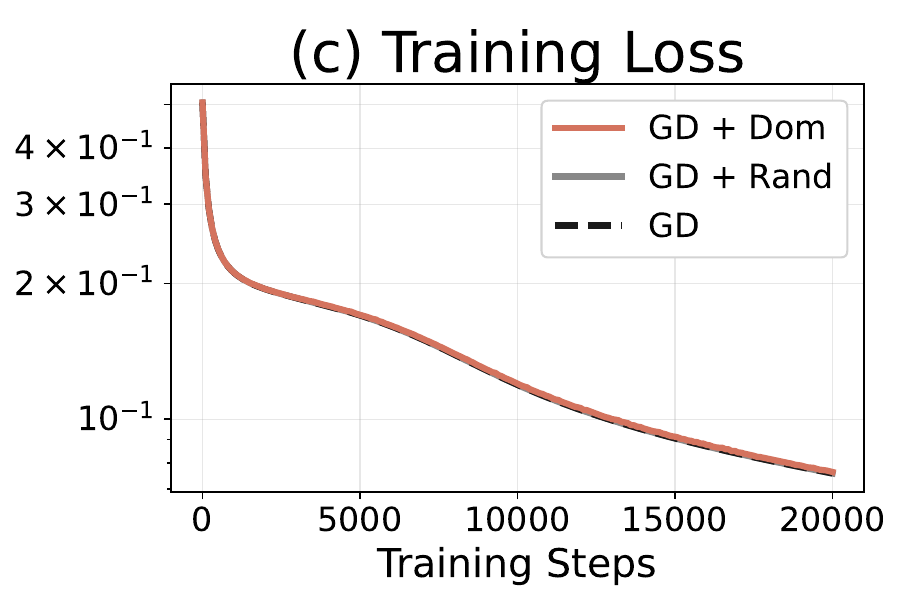}
      \vspace{-1mm} 
    \caption{
\textbf{Dominant perturbations reduce top-\(k\) sharpness.}
We compare GD, GD with dominant mean-zero perturbations (GD+Dom), and GD with random mean-zero perturbations (GD+Rand).
The two perturbation runs are matched in rank and noise scale \((k=10,\rho=0.1)\).
(a) Dominant perturbations induce strong dominant alignment.
(b) Only dominant perturbations reduce \(S_k\).
(c) Both perturbed runs leave the GD loss curve nearly unchanged.
All experiments are run on an MLP.
}
\label{fig:perturbation}
  \end{figure}

 \vspace{-1mm}


The previous observation showed that the dominant subspace is related to sharpness reduction rather than loss reduction.
However, this simple comparison between Dom-SGD and Bulk-SGD does not reveal whether this sharpness reduction comes from the \textbf{gradient component in the dominant
subspace} or from \textbf{stochastic motion within that subspace}.

To separate these effects, we conduct a controlled noise injection experiment that keeps the mean update equal to full-batch GD while adding only mean-zero perturbations in different subspaces. Specifically, we compare the perturbed GD dynamics \(\theta_{t+1}=\theta_t-\eta(\nabla L(\theta_t)+\zeta_t)\) with \(\zeta_t\) sampled from different subspaces.
  For the dominant perturbation, we set \(\zeta_t\sim\mathcal N(0,\rho^2 P_{\mathrm{dom}}(\theta_t))\), and for the random perturbation, we set \(\zeta_t\sim\mathcal N(0,\rho^2 P_{\mathrm{rand}})\),
  where \(P_{\mathrm{rand}}\) is a random \(k\)-dimensional orthogonal projection fixed at initialization.

First, as shown in Figure~\ref{fig:perturbation}(c), GD, GD with random perturbations (GD+Rand), and GD with dominant perturbations (GD+Dom) follow almost the same training-loss curve, indicating that these perturbations do not significantly alter the rate of loss reduction. However, as shown in Figure~\ref{fig:perturbation}(a) and ~\ref{fig:perturbation}(b), their trajectories exhibit substantial geometric differences: GD and GD+Rand maintain low \(\chi_k\) and high \(S_k\), whereas GD+Dom drives \(\chi_k\) close to 1 and maintains a lower top-\(k\) sharpness \(S_k\) than either GD or GD+Rand. 

These results show two things. First, as suggested by the previous observation, alignment with the dominant subspace does not imply that loss reduction occurs along the dominant directions. Second, and more importantly, stochastic motion alone within the dominant subspace can actually affect sharpness dynamics while barely changing the loss curve. Together, these results indicate that the role of the dominant subspace lies not in providing a direction for loss reduction, but rather in changing the \textbf{local geometry associated with the dominant directions}, especially top-\(k\) sharpness. In the next section, we show that this effect arises from the interaction between the covariance of a mean-zero displacement and local variations of the Hessian.


 \vspace{-3mm}

\vspace{-1mm}

\section{Sharpness Correction Term}
\label{sec:correction}
    \vspace{-1mm}
In this section, we show that when the parameter fluctuates in the dominant directions around a reference point, averaging the gradient leaves a \textbf{\textit{sharpness correction term}}.
To show this, we consider a displacement \(\delta\in E_{\mathrm{dom}}(\theta_c)\) from a reference point \(\theta_c\), with \(\mathbb E[\delta]=0\), and write the nearby parameter as \(\theta_c+\delta\).
Expanding \(\nabla L(\theta_c+\delta)\) using a Taylor expansion around \(\theta_c\), we get:
\begin{equation}
\nabla L(\theta_c+\delta)
=
\nabla L(\theta_c)
+
H(\theta_c)\delta
+
\frac12
\nabla(\delta^\top H\delta)(\theta_c)
+
O(\|\delta\|^3).
\label{eq:taylor_individual}
\end{equation}
The first term in Eq.~\eqref{eq:taylor_individual} is the gradient at the reference point, the second term is the linear term in the displacement, and the third term is a second-order term that captures the local variation of the Hessian. Averaging this over \(\delta\) gives:
\vspace{-2mm}
\begin{equation}
\mathbb E_\delta[\nabla L(\theta_c+\delta)]
=
\nabla L(\theta_c)
+
\underbrace{H(\theta_c)\mathbb E[\delta]}_{=\,0}
+
\frac12
\nabla \operatorname{Tr}\!\left(H(\theta_c)C\right)
+
O(\mathbb E\|\delta\|^3),
\qquad
C:=\mathbb E[\delta\delta^\top].
\label{eq:taylor_average}
\end{equation}
Here, \(C\) is held fixed when taking the derivative.
As shown in Eq.~\eqref{eq:taylor_average}, the averaged gradient is the sum of the gradient at the reference point and an additional term $\frac12
\nabla \operatorname{Tr}\!\left(H(\theta_c)C\right)$.
This term is determined by the displacement covariance \(C\) and acts to lower the covariance-weighted curvature represented by \(\operatorname{Tr}(HC)\).
When \(C\) is supported in the dominant subspace, it lowers the weighted curvature associated with top-\(k\) sharpness, so we refer to it as a \textit{sharpness correction term}.

\paragraph{Mini-batch-induced displacement covariance.}
Finally, we derive the displacement covariance induced by mini-batch noise in the local recursion and substitute it into the sharpness correction term. This shows that the sharpness effect of mini-batch noise can be represented as a deterministic correction term.

\begingroup
\looseness=-1
To define the local displacement created by mini-batch noise around a reference point \(\theta_c\), we compare the mini-batch SGD update with the full-batch GD update.
Consider a local state \(\theta_c+\delta_s\), where \(\delta_s\in E_{\mathrm{dom}}(\theta_c)\), and let \(P_c:=P_{\mathrm{dom}}(\theta_c)\).
The next displacement \(\delta_{s+1}\) is defined as
\endgroup
\[
\delta_{s+1}
:=
P_c(\theta_{\mathrm{mb}}^+-\theta_{\mathrm{gd}}^+),
\qquad
\theta_{\mathrm{mb}}^+
:=
\theta_c+\delta_s-\eta\nabla L_{B_s}(\theta_c+\delta_s),
\qquad
\theta_{\mathrm{gd}}^+
:=
\theta_c-\eta\nabla L(\theta_c).
\]
We then linearize the mini-batch gradient at \(\theta_c+\delta_s\) around \(\theta_c\):
\[
\nabla L_{B_s}(\theta_c+\delta_s)
\approx
\nabla L(\theta_c)+H_c\delta_s+\xi_{B_s}(\theta_c),
\qquad
H_c:=H(\theta_c).
\]
Substituting this into the definition of \(\delta_{s+1}\) gives
\begin{equation}
\delta_{s+1}
=
P_c\bigl(\delta_s-\eta H_c\delta_s-\eta\xi_{B_s}(\theta_c)\bigr)
=
A_c\delta_s-\eta P_c\xi_{B_s}(\theta_c),
\qquad
A_c:=P_c(I-\eta H_c)P_c.
\label{eq:local_displacement_recursion}
\end{equation}
Taking second moments in Eq.~\eqref{eq:local_displacement_recursion} gives
\[
C_{s+1}
=
A_cC_sA_c^\top+\eta^2\Sigma_{\mathrm{dom}}(\theta_c),
\qquad
(C_s:=\mathbb E[\delta_s\delta_s^\top]),
\quad
(\Sigma_{\mathrm{dom}}(\theta_c):=P_c\operatorname{Cov}(\xi_B(\theta_c))P_c).
\]
Here, \(\Sigma_{\mathrm{dom}}\) denotes the mini-batch noise covariance projected onto the dominant subspace. Let \(C_{\mathrm{mb}}\) be the stationary limit of the covariance sequence \(C_s\). Then
\(C_{\mathrm{mb}}=A_cC_{\mathrm{mb}}A_c^\top+\eta^2\Sigma_{\mathrm{dom}}\).
Therefore, substituting \(C_{\mathrm{mb}}\) into Eq.~\eqref{eq:taylor_average} gives the following:

\vspace{-4mm}

\[
\begin{aligned}
\mathbb{E}[\nabla L(\theta_c+\delta)]
\approx
\nabla L(\theta_c)
+
\frac{1}{2}
\nabla \operatorname{Tr}\!\left(H(\theta_c)C_{\mathrm{mb}}\right),
\qquad
C_{\mathrm{mb}}
=
\eta^2
\sum_{\ell=0}^{\infty}
A_c^\ell
\Sigma_{\mathrm{dom}}
(A_c^\top)^\ell .
\end{aligned}
\]

Thus, \(C_{\mathrm{mb}}\) is not merely the mini-batch noise covariance, but the displacement covariance produced by the local recursion, and therefore it also depends on the local Hessian structure and the learning rate.
With this covariance, the effect of mini-batch noise on the averaged gradient is represented as a deterministic correction term
(see Appendix~\ref{app:derivation} for the detailed derivation).

\vspace{-3mm}
\section{Experiments}
\label{sec:experiments}

\begin{figure}[!t]
  \centering
  \includegraphics[width=0.32\textwidth]{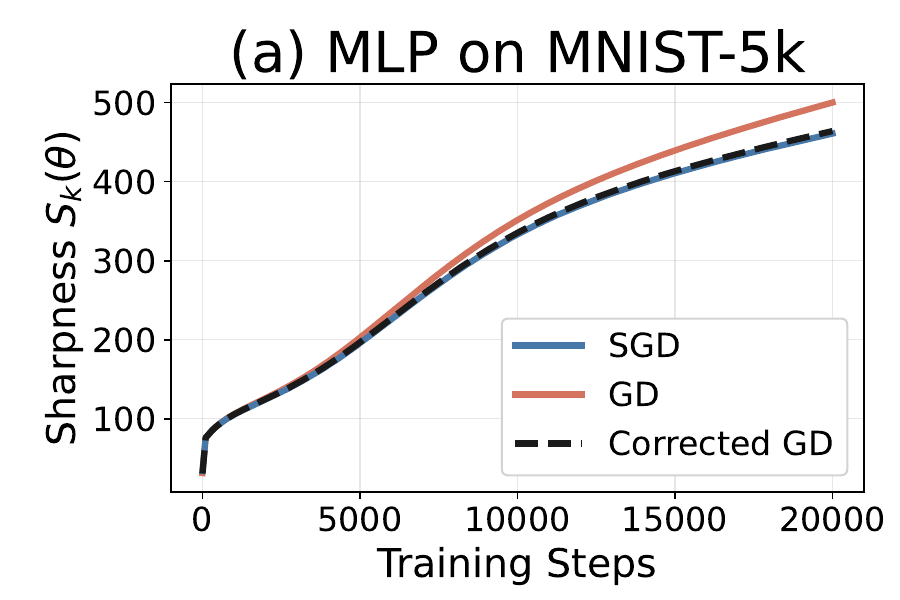}\hfill
  \includegraphics[width=0.32\textwidth]{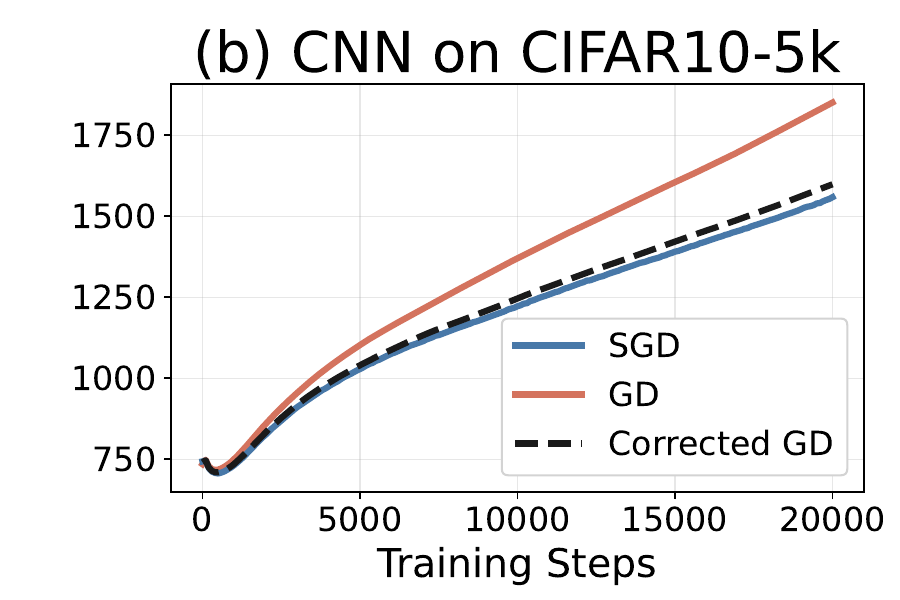}\hfill
  \includegraphics[width=0.32\textwidth]{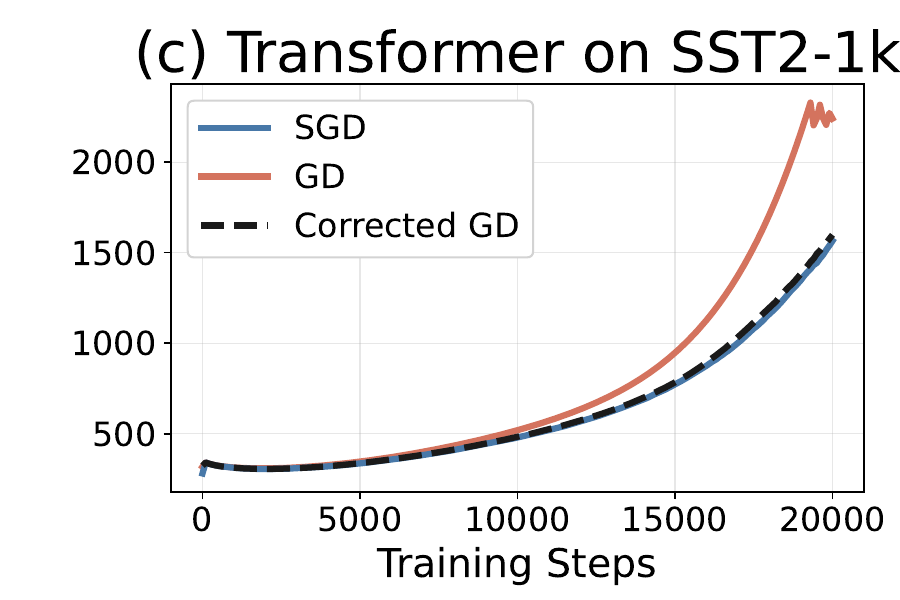}
  \caption{
\textbf{The sharpness correction term brings GD closer to mini-batch SGD in top-\(k\) sharpness.}
Across MLP, CNN, and Transformer, full-batch GD reaches higher top-\(k\) sharpness \(S_k\),
whereas adding the derived sharpness correction term to GD brings its
top-\(k\) sharpness curve closer to that of mini-batch SGD.}
  \label{fig:teaser}
\end{figure}

In this section, we empirically test whether the sharpness correction term derived in Section~\ref{sec:correction} reproduces the low sharpness of mini-batch SGD. 
For the experiments, we compare the sharpness dynamics of full-batch GD, mini-batch SGD, and corrected GD across three model architectures: MLP, CNN, and Transformer~\citep{vaswani2017attention}.
Here, corrected GD is obtained by adding our sharpness correction term to GD,
\(\theta_{t+1}=\theta_t-\eta\nabla L(\theta_t)+\eta b_{\mathrm{corr}}(\theta_t)\),
where \(b_{\mathrm{corr}}(\theta):=-\frac12\nabla_{\vartheta}\operatorname{Tr}\!\left(H(\vartheta)C_{\mathrm{mb}}(\theta)\right)\big|_{\vartheta=\theta}\).
Detailed experimental settings are given in Appendix~\ref{app:experimental_setup}.

Figure~\ref{fig:teaser} shows that full-batch GD moves toward higher
\(S_k\) than mini-batch SGD in all three models, whereas corrected GD stays much closer to the low sharpness level of SGD.
In particular, for the MLP, corrected GD almost overlaps with SGD, and for the CNN and Transformer, it also keeps \(S_k\) lower than GD and substantially reduces the \textit{sharpness gap} with SGD.

This result shows that the sharpness correction term we derived in Section \ref{sec:correction} explains much of the sharpness dynamics of SGD, empirically showing that the dominant subspace helps explain the lower top-\(k\) sharpness of mini-batch SGD, rather than directly providing a direction for reducing the loss.

\vspace{-3mm}

\section{Conclusion}

Prior work by \citet{song2025does} shows that the dominant subspace, the top-\(k\) eigenspace of the loss Hessian, does not provide meaningful directions for loss reduction, leaving its explanatory role unclear. In this paper, we show that this dominant subspace plays an important role in reducing top-\(k\) sharpness under SGD. Specifically, we show that perturbations within the dominant subspace reduce top-\(k\) sharpness, and derive the sharpness correction term induced by mini-batch noise. Empirically, we show that adding this correction term to GD brings its sharpness evolution closer to that of mini-batch SGD, demonstrating that the dominant subspace matters not as a direction for loss reduction, but as the subspace through which mini-batch noise lowers sharpness.

\clearpage
\bibliography{sample}


\clearpage
\appendix

\makeatletter
\@removefromreset{figure}{section}
\makeatother
\renewcommand{\thefigure}{\arabic{figure}}
\setcounter{figure}{4}

\startcontents[appendix]

\phantomsection
{\centering\Large\bfseries Appendix\par}
\vspace{4mm}

\printcontents[appendix]{}{1}{}
\clearpage

\section{Related Work}
\label{app:related}

\paragraph{Hessian spectra and dominant alignment.}
The training-loss Hessian of a neural network often has a structure that separates into a small number of large outlier eigenvalues and a broad bulk spectrum
\citep{sagun2017eigenvalues,DBLP:journals/corr/SagunEGDB17,pmlr-v97-ghorbani19b,pmlr-v97-papyan19a,JMLR:v21:20-933}. It has also been repeatedly observed that, during training, the training-loss gradient is strongly aligned with a low-dimensional subspace spanned by a small number of eigenvectors corresponding to the large eigenvalues of the Hessian
\citep{gurari2018gradientdescenthappenstiny,pmlr-v97-ghorbani19b}.
These observations suggest that SGD dynamics in a high-dimensional parameter space are closely related to low-dimensional structure.

\paragraph{Projected-update experiments.}
\citet{song2025does} conducted projected-update experiments to test whether SGD can still reduce the training loss when its updates are restricted to the top Hessian eigenspace, referred to as the dominant subspace. Surprisingly, Dom-SGD, which trains using only the SGD update projected onto the dominant subspace, barely reduces the training loss. In contrast, Bulk-SGD, which trains using the SGD update with its dominant component removed, continues to reduce the loss similarly to standard SGD.
To address this phenomenon, \citet{song2025does} interpret the loss landscape as an ill-conditioned valley. In this view, the dominant directions correspond to high-curvature valley walls, so Dom-SGD mostly moves across the walls and has difficulty reducing the loss, whereas Bulk-SGD follows the valley floor and continues to train. This intuition is also similar to the river-valley loss landscape \citep{wen2025understanding}. \citet{deng2026suspiciousalignmentsgdfinegrained} further analyze projected SGD updates by deriving step-size conditions for loss decrease. However, these loss-based explanations account for why a dominant-projected update has difficulty making loss progress, but not why the dominant subspace is connected to lower top-\(k\) sharpness. This question is the starting point of our work.

\paragraph{Sharpness and flatness.}
Sharpness and flatness have long been studied as loss-landscape geometry for explaining generalization. Early work suggested that flat minima, where the loss does not change much under small perturbations around the solution, are associated with better generalization
\citep{10.1162/neco.1997.9.1.1}. Later, it was also observed that large-batch training can converge to sharper minimizers than small-batch training and show a generalization gap, with worse test performance despite similar training loss~\citep{keskar2017on}. This view also led to optimization methods that directly search for flat minima. A representative method is SAM, and since then, related variants and theoretical analyses have also been actively studied
\citep{foret2021sharpnessaware,pmlr-v139-kwon21b,pmlr-v162-andriushchenko22a,JMLR:v24:23-043,wen2023how}. On the other hand, since sharpness-based measures can be sensitive to parameterization or experimental settings, there is also a view that sharpness alone has limits in explaining generalization
\citep{pmlr-v70-dinh17b,Jiang*2020Fantastic}.

\paragraph{SGD noise and implicit regularization.}
SGD noise has been studied as a source of implicit regularization related to flatness and sharpness. Prior work has studied SGD noise through the stationary distribution of constant-step SGD \citep{mandt2018stochasticgradientdescentapproximate}, or approximated SGD using SDEs or stochastic modified equations
\citep{pmlr-v70-li17f,JMLR:v20:17-526,li2021on}. Along with this view, other studies explored how the scale and covariance structure of SGD noise affect optimization and generalization
\citep{jastrzębski2018three,pmlr-v97-zhu19e,pmlr-v108-thomas20a,pmlr-v108-wen20a}. In particular, some analyses suggest that this covariance structure is related to sharp directions and affects flat-minima selection \citep{NEURIPS2022_1e55c38d}. There is also work showing that mini-batch SGD noise near a zero-loss manifold induces a drift toward lower sharpness of the loss \citep{li2022what}. Like this line of work, we study how SGD noise affects sharpness, but we focus on the displacement covariance induced in the dominant subspace rather than on the mini-batch noise covariance itself.

\paragraph{Edge of stability.}
In large-step-size GD, the largest Hessian eigenvalue is observed to increase up to around the stability threshold \(2/\eta\), after which training continues without diverging, although the loss becomes non-monotonic. This phenomenon is called the Edge of Stability (EoS)
\citep{cohen2021gradient}.
Subsequent theoretical analyses showed that motion caused by instability in the top Hessian direction can lower sharpness through higher-order terms
\citep{damian2023selfstabilizationimplicitbiasgradient},
and that the time-averaged trajectory of large-step GD near EoS can be described by a deterministic flow
\citep{cohen2025understanding}. Our mechanism is closely related to these analyses in that motion in high-curvature directions leaves a higher-order correction after averaging.
However, the source of the correction is different.
While these works study oscillations that arise in large-step GD, we focus on fluctuations induced by mini-batch noise in SGD.

\section{Notation and Assumptions}
\label{app:notation}
In this section, we fix the notation and assumptions used in Appendix~\ref{app:derivation}.

\subsection{Notation}
\label{app:notation_basic}
\paragraph{Optimization setup.}
We use \(L:\mathbb R^d\to\mathbb R\) to denote the training loss. For a mini-batch \(B\), we write \(L_B\) for the corresponding mini-batch loss.

We write
\[
H(\theta):=\nabla^2 L(\theta)
\]
for the Hessian matrix at \(\theta\).
For a mini-batch \(B\), we define the mini-batch noise by
\[
\xi_B(\theta):=\nabla L_B(\theta)-\nabla L(\theta),
\qquad
\mathbb E_B[\xi_B(\theta)\mid \theta]=0.
\]
We denote the corresponding noise covariance by
\[
\Sigma_{\xi}(\theta)
:=
\operatorname{Cov}_B(\xi_B(\theta)\mid \theta).
\]
Mini-batch SGD with learning rate \(\eta>0\) is written as
\[
\theta_{t+1}
=
\theta_t-\eta\nabla L_{B_t}(\theta_t)
=
\theta_t-\eta\bigl(\nabla L(\theta_t)+\xi_{B_t}(\theta_t)\bigr).
\]

\paragraph{Dominant Hessian subspace.}

We denote the eigenpairs of \(H(\theta)\) by
\((\lambda_i(\theta),e_i(\theta))_{i=1}^d\), ordered as
\[
\lambda_1(\theta)\ge \lambda_2(\theta)\ge \cdots \ge \lambda_d(\theta).
\]
We define the dominant subspace as
\[
E_{\mathrm{dom}}(\theta)
:=
\operatorname{span}\{e_1(\theta),\ldots,e_k(\theta)\},
\]
and write \(P_{\mathrm{dom}}(\theta)\) for the orthogonal projection onto
\(E_{\mathrm{dom}}(\theta)\).
We also write
\[
E_{\mathrm{bulk}}(\theta):=E_{\mathrm{dom}}(\theta)^\perp,
\qquad
P_{\mathrm{bulk}}(\theta):=I-P_{\mathrm{dom}}(\theta).
\]
We define the top-\(k\) sharpness as
\[
S_k(\theta):=\sum_{i=1}^k \lambda_i(\theta).
\]

\paragraph{Local displacements.}
For a reference point \(\theta_c\), we write a nearby parameter as
\[
\theta_c+\delta,
\]
where \(\delta\) denotes the local displacement.
In the local averaging calculations, \(\delta\) is mean-zero, and we write
\[
C(\theta_c)=\mathbb E[\delta\delta^\top]
\]
for its displacement covariance.
When the displacement is supported on \(E_{\mathrm{dom}}(\theta_c)\), the covariance \(C\) is supported on the same subspace.

\paragraph{Local linear map.}
For a reference point \(\theta_c\), we define 
\[
A_c
:=
P_{\mathrm{dom}}(\theta_c)
\bigl(I-\eta H(\theta_c)\bigr)
P_{\mathrm{dom}}(\theta_c),
\]
viewed as a linear map on \(E_{\mathrm{dom}}(\theta_c)\).
This is the linear part of the local recursion derived in Appendix~\ref{app:mb_covariance}.

\subsection{Local Assumptions}
\label{app:local_assumptions}

We use the following local assumptions at each reference point \(\theta_c\).

\paragraph{Assumption 1 (Local Taylor regularity).}
Around \(\theta_c\), the loss \(L\) is locally \(C^4\).
This is used to apply the Taylor expansion of the gradient and to control the local remainder terms.

\paragraph{Assumption 2 (Eigengap).}
When derivatives of \(S_k\) or \(P_{\mathrm{dom}}\) are used, we assume
\[
\lambda_k(\theta_c)>\lambda_{k+1}(\theta_c).
\]
This ensures that the top-\(k\) eigenspace is separated from the rest of the spectrum near \(\theta_c\).

\paragraph{Assumption 3 (Moments and mini-batch noise).}

We use the following moment and noise assumptions.

\vspace{5mm}

\begin{description}[
  style=nextline,
  leftmargin=1.5em,
  labelindent=0pt,
  font=\normalfont\bfseries,
  itemsep=0.4em,
  topsep=0.2em
]

\textbf{(i) Local displacement moments.}
The local displacements used in the averaging calculations satisfy
\[
\mathbb E[\delta]=0,
\qquad
\mathbb E\|\delta\|^3<\infty.
\]
In the local recursion, we initialize with
\[
\mathbb E[\delta_0]=0.
\]

\textbf{(ii) Mini-batch noise.}
The mini-batch noise satisfies
\[
\mathbb E_B[\xi_B(\theta_c)\mid \theta_c]=0,
\qquad
\|\Sigma_{\xi}(\theta_c)\|<\infty.
\]

\textbf{(iii) Local noise freezing.}
For the local recursion, we write the noise variation as
\[
\xi_{B_s}(\theta_c+\delta_s)
=
\xi_{B_s}(\theta_c)+r_{\xi,s},
\]
and ignore \(r_{\xi,s}\) when linearizing the mini-batch gradient around \(\theta_c\).

\makeatletter\@newlistfalse\makeatother
\end{description}

\paragraph{Assumption 4 (Local stability).}
When taking the stationary limit of the covariance recursion, we assume that the selected dominant eigenvalues satisfy

\[
0<\lambda_k(\theta_c)\le \lambda_1(\theta_c)<\frac{2}{\eta}.
\]
Since the local linear map \(A_c\) has eigenvalues
\(1-\eta\lambda_i(\theta_c)\) on \(E_{\mathrm{dom}}(\theta_c)\), this implies
\[
\rho(A_c)<1.
\]

\section{Derivation of the Sharpness Correction Term}
\label{app:derivation}

In this section, we derive in more detail the sharpness correction term presented briefly in the main text.
Section~\ref{sec:correction} showed that when the local displacement lies in the dominant directions and has zero mean, averaging the gradient leaves a sharpness correction term, and described how this correction is induced by mini-batch noise.
Here, we give the same calculation in more detail and show how the dominant perturbation experiment in Section~\ref{sec:main} and mini-batch noise each lead to a sharpness correction term through their displacement covariance.
The notation and assumptions follow Appendix~\ref{app:notation}.

First, in Appendix~\ref{app:avg_gradient}, we show that averaging \(\nabla L(\theta_c+\delta)\) over a local displacement with zero mean leaves a covariance-weighted curvature term: the sharpness correction term.
Next, in Appendix~\ref{app:dominant_cases}, we specialize this sharpness correction term to the dominant perturbation experiment in Section~\ref{sec:main}, and show why it lowers top-\(k\) sharpness in that case.
Finally, in Appendix~\ref{app:mb_covariance}, we compute the displacement covariance induced by mini-batch noise through a local recursion, and show how it determines the mini-batch-induced sharpness correction term.

\subsection{Sharpness correction from local displacement}
\label{app:avg_gradient}

We compute the covariance-weighted curvature term that remains when we average the gradient around a reference point.
To do this, fix a reference point \(\theta_c\), and let \(\delta\) be a local displacement around \(\theta_c\) with \(\mathbb E[\delta]=0\).
We write the nearby parameter as \(\theta_c+\delta\), and define the displacement covariance by
\[
C(\theta_c):=\mathbb E[\delta\delta^\top].
\]

Motivated by the dominant perturbation experiment in Section~\ref{sec:main}, we consider the case where the displacement around \(\theta_c\) lies in the dominant subspace.
That is, we take \(\delta\in E_{\mathrm{dom}}(\theta_c)\), and so we view \(C\) as a covariance supported on
\(E_{\mathrm{dom}}(\theta_c)\).

Now expand the gradient at \(\theta_c+\delta\) around \(\theta_c\).
The second-order Taylor expansion of the gradient is
\begin{equation}
\nabla L(\theta_c+\delta)
=
\nabla L(\theta_c)
+
H(\theta_c)\delta
+
\frac12\nabla(\delta^\top H\delta)(\theta_c)
+
R_3(\theta_c,\delta).
\label{eq:app_gradient_taylor}
\end{equation}
In Eq.~\eqref{eq:app_gradient_taylor}, the first term is the gradient at the reference point, and the second term is the linear term in the displacement.
The third term is a second-order term that captures the local variation of the Hessian around \(\theta_c\), and \(R_3(\theta_c,\delta)\) is the higher-order remainder.

Now average Eq.~\eqref{eq:app_gradient_taylor} over \(\delta\).
Since \(\mathbb E[\delta]=0\), the linear term vanishes after averaging:
\[
\mathbb E_\delta[H(\theta_c)\delta]
=
H(\theta_c)\mathbb E_\delta[\delta]
=
0.
\]
After this cancellation, the only nontrivial term left to average is the second-order term containing the Hessian variation.
To compute this, let \(\vartheta\) be a temporary variable and write
\[
\delta^\top H(\vartheta)\delta
=
\operatorname{Tr}\!\left(H(\vartheta)\delta\delta^\top\right).
\]
Then, since \(C(\theta_c)=\mathbb E[\delta\delta^\top]\),
\[
\mathbb E_\delta[\delta^\top H(\vartheta)\delta]
=
\operatorname{Tr}\!\left(H(\vartheta)C(\theta_c)\right).
\]
Here, \(C(\theta_c)\) is held fixed when taking the derivative with respect to \(\vartheta\).
Therefore,
\[
\mathbb E_\delta
\left[
\nabla(\delta^\top H\delta)(\theta_c)
\right]
=
\left.
\nabla_{\vartheta}
\operatorname{Tr}\!\left(H(\vartheta)C(\theta_c)\right)
\right|_{\vartheta=\theta_c}.
\]
Thus, the averaged gradient is
\begin{equation}
\mathbb E_\delta[\nabla L(\theta_c+\delta)]
=
\nabla L(\theta_c)
+
\frac12
\left.
\nabla_{\vartheta}
\operatorname{Tr}\!\left(H(\vartheta)C(\theta_c)\right)
\right|_{\vartheta=\theta_c}
+
\mathcal R_3(\theta_c),
\label{eq:app_avg_gradient_expansion}
\end{equation}
where
\[
\mathcal R_3(\theta_c)
:=
\mathbb E_\delta[R_3(\theta_c,\delta)].
\]
Under Assumptions 1 and 3 in Appendix~\ref{app:local_assumptions}, this remainder is \(O(\mathbb E_\delta\|\delta\|^3)\).

As shown in Eq.~\eqref{eq:app_avg_gradient_expansion}, averaging the gradient around the reference point adds a covariance-weighted curvature term to the gradient at the reference point.
We refer to this additional term as a sharpness correction term.
This term is determined once the covariance \(C\) is specified.
We first consider the covariance that corresponds to the dominant perturbation experiment in Section~\ref{sec:main}, and then derive the covariance created by mini-batch noise.

\subsection{Sharpness correction from dominant perturbations}
\label{app:dominant_cases}

For the dominant perturbation experiment in Section~\ref{sec:main}, we specify the covariance that enters the sharpness correction term in Eq.~\eqref{eq:app_avg_gradient_expansion}.
We then show that the resulting correction lowers top-\(k\) sharpness.

A dominant perturbation can be viewed as the case where the covariance is isotropic inside the dominant subspace.
That is, for some scalar variance \(\sigma^2>0\), let
\begin{equation}
C=\sigma^2P_{\mathrm{dom}}(\theta_c)
\label{eq:app_dom_iso_cov}
\end{equation}

We substitute the covariance in Eq.~\eqref{eq:app_dom_iso_cov} into the correction term in
Eq.~\eqref{eq:app_avg_gradient_expansion}.
Under Assumption 2 in Appendix~\ref{app:local_assumptions}, \(S_k\) is locally differentiable, and the standard eigenvalue derivative formula gives
\[
\left.
\nabla_{\vartheta}
\operatorname{Tr}\!\left(H(\vartheta)P_{\mathrm{dom}}(\theta_c)\right)
\right|_{\vartheta=\theta_c}
=
\nabla S_k(\theta_c).
\]
Therefore, by Eq.~\eqref{eq:app_dom_iso_cov},
\begin{equation}
\left.
\nabla_{\vartheta}
\operatorname{Tr}\!\left(H(\vartheta)C\right)
\right|_{\vartheta=\theta_c}
=
\sigma^2\nabla S_k(\theta_c).
\label{eq:app_dom_grad_sk}
\end{equation}
Substituting this into Eq.~\eqref{eq:app_avg_gradient_expansion} gives
\begin{equation}
\mathbb E_\delta[\nabla L(\theta_c+\delta)]
=
\nabla L(\theta_c)
+
\frac{\sigma^2}{2}\nabla S_k(\theta_c)
+
\mathcal R_3(\theta_c).
\label{eq:app_dom_avg_gradient}
\end{equation}
As shown in Eq.~\eqref{eq:app_dom_avg_gradient}, when the covariance is isotropic inside the dominant subspace, the averaged gradient has the form of the gradient at the reference point plus the term
\(\frac{\sigma^2}{2}\nabla S_k(\theta_c)\).
Thus, in the negative-gradient update, the term
\(-\frac{\sigma^2}{2}\nabla S_k(\theta_c)\) is added on top of the full-gradient term, and this term acts in the direction that lowers \(S_k\).
This explains why the dominant perturbation in Section~\ref{sec:main}
shows lower top-\(k\) sharpness than GD.

\subsection{Sharpness correction from mini-batch noise}
\label{app:mb_covariance}

In Appendix~\ref{app:dominant_cases}, we considered the case where the displacement covariance can be specified directly, as in the dominant perturbation experiment in Section~\ref{sec:main}.
For mini-batch noise, this covariance is not directly given.
We now derive it from a local recursion around a reference point.

\paragraph{Local recursion.}
Fix a reference point \(\theta_c\). To model the local displacement created by mini-batch noise around this point, we introduce a local recursion around \(\theta_c\), indexed by \(s\).
At local step \(s\), we write the local state as \(\theta_c+\delta_s\), with \(\delta_s\in E_{\mathrm{dom}}(\theta_c)\), and draw a fresh mini-batch \(B_s\).

To define one step of this local recursion, we take the difference between the mini-batch update from \(\theta_c+\delta_s\) and the full-batch GD update from \(\theta_c\), and project it onto the dominant subspace.
First, define the mini-batch update at \(\theta_c+\delta_s\) and the full-batch GD update at \(\theta_c\) as
\begin{equation}
\theta_{\mathrm{mb}}^+
:=
\theta_c+\delta_s-\eta\nabla L_{B_s}(\theta_c+\delta_s),
\label{eq:app_mb_update}
\end{equation}
\begin{equation}
\theta_{\mathrm{gd}}^+
:=
\theta_c-\eta\nabla L(\theta_c)
\label{eq:app_gd_update}
\end{equation}
respectively.
Then the next displacement in the local recursion is defined as
\begin{equation}
\delta_{s+1}
:=
P_{\mathrm{dom}}(\theta_c)(\theta_{\mathrm{mb}}^+-\theta_{\mathrm{gd}}^+).
\label{eq:app_local_update_difference}
\end{equation}
That is, after subtracting the full-batch GD step from \(\theta_c\), we project the remaining local difference onto the dominant subspace and use the result as \(\delta_{s+1}\).

Next, we approximate the mini-batch gradient
\(\nabla L_{B_s}(\theta_c+\delta_s)\) in Eq.~\eqref{eq:app_mb_update} around the reference point \(\theta_c\).
By the definition of mini-batch noise,
\(\xi_B(\theta):=\nabla L_B(\theta)-\nabla L(\theta)\), we have
\begin{equation}
\nabla L_{B_s}(\theta_c+\delta_s)
=
\nabla L(\theta_c+\delta_s)
+
\xi_{B_s}(\theta_c+\delta_s).
\label{eq:app_mb_noise_decomposition}
\end{equation}
Now approximate the two terms in Eq.~\eqref{eq:app_mb_noise_decomposition} around \(\theta_c\) as follows:
\[
\nabla L(\theta_c+\delta_s)
=
\nabla L(\theta_c)
+
H(\theta_c)\delta_s
+
O(\|\delta_s\|^2),
\qquad
\xi_{B_s}(\theta_c+\delta_s)
\approx
\xi_{B_s}(\theta_c).
\]
Therefore, after substituting this local approximation into Eq.~\eqref{eq:app_mb_noise_decomposition} and dropping the higher-order term, we get
\begin{equation}
\nabla L_{B_s}(\theta_c+\delta_s)
\approx
\nabla L(\theta_c)
+
H(\theta_c)\delta_s
+
\xi_{B_s}(\theta_c).
\label{eq:app_mb_gradient_linearization}
\end{equation}
Eq.~\eqref{eq:app_mb_gradient_linearization} shows that the mini-batch gradient at
\(\theta_c+\delta_s\) is approximated by the full-gradient term at \(\theta_c\), the linear response to the local displacement, and the mini-batch noise term at \(\theta_c\).

Now use Eq.~\eqref{eq:app_mb_gradient_linearization} to compute
\(\delta_{s+1}\).
By Eq.~\eqref{eq:app_mb_update} and Eq.~\eqref{eq:app_local_update_difference},
\begin{equation}
\begin{aligned}
\delta_{s+1}
&\approx
P_{\mathrm{dom}}(\theta_c)
\Big[
\theta_c+\delta_s
-\eta\bigl(
\nabla L(\theta_c)
+
H(\theta_c)\delta_s
+
\xi_{B_s}(\theta_c)
\bigr)
-
\bigl(\theta_c-\eta\nabla L(\theta_c)\bigr)
\Big] \\
&=
P_{\mathrm{dom}}(\theta_c)
\bigl(
\delta_s-\eta H(\theta_c)\delta_s-\eta\xi_{B_s}(\theta_c)
\bigr) \\
&=
A_c\delta_s
-
\eta P_{\mathrm{dom}}(\theta_c)\xi_{B_s}(\theta_c).
\end{aligned}
\label{eq:app_dom_displacement_recursion}
\end{equation}
Eq.~\eqref{eq:app_dom_displacement_recursion} is the local recursion around the reference point \(\theta_c\).
Here, \(A_c\delta_s\) is the term by which the previous local displacement is carried forward through \(A_c\), and
\(-\eta P_{\mathrm{dom}}(\theta_c)\xi_{B_s}(\theta_c)\) is the term by which the current mini-batch noise is newly injected into the dominant subspace.

\paragraph{Displacement covariance recursion.}
Eq.~\eqref{eq:app_dom_displacement_recursion} gives the recursion for the local displacement \(\delta_s\).
Now we compute the local displacement covariance induced by this recursion.
First, define the \(s\)-th local displacement covariance as
\begin{equation}
C_s:=\mathbb E[\delta_s\delta_s^\top].
\label{eq:app_local_displacement_covariance}
\end{equation}
Next, define the dominant noise covariance at the reference point \(\theta_c\) as
\begin{equation}
\Sigma_{\mathrm{dom}}(\theta_c)
:=
P_{\mathrm{dom}}(\theta_c)
\Sigma_{\xi}(\theta_c)
P_{\mathrm{dom}}(\theta_c).
\label{eq:app_dom_noise_covariance}
\end{equation}

Since each \(B_s\) is a mini-batch drawn freshly at the local step,
\(\xi_{B_s}(\theta_c)\) is independent of \(\delta_s\), and
\(\mathbb E[\xi_{B_s}(\theta_c)\mid\theta_c]=0\).
From these conditions, the cross terms satisfy
\[
\mathbb E[\delta_s\xi_{B_s}(\theta_c)^\top]
=
\mathbb E\!\left[
\delta_s\,
\mathbb E[\xi_{B_s}(\theta_c)^\top\mid \delta_s,\theta_c]
\right]
=
0,
\qquad
\mathbb E[\xi_{B_s}(\theta_c)\delta_s^\top]=0.
\]
Therefore, computing \(C_{s+1}=\mathbb E[\delta_{s+1}\delta_{s+1}^\top]\) from Eq.~\eqref{eq:app_dom_displacement_recursion} gives
\[
\begin{aligned}
C_{s+1}
&=
\mathbb E[\delta_{s+1}\delta_{s+1}^\top] \\
&=
\mathbb E\!\left[
\left(
A_c\delta_s-\eta P_{\mathrm{dom}}(\theta_c)\xi_{B_s}(\theta_c)
\right)
\left(
A_c\delta_s-\eta P_{\mathrm{dom}}(\theta_c)\xi_{B_s}(\theta_c)
\right)^\top
\right] \\
&=
A_cC_sA_c^\top
+
\eta^2
P_{\mathrm{dom}}(\theta_c)
\Sigma_{\xi}(\theta_c)
P_{\mathrm{dom}}(\theta_c).
\end{aligned}
\]
Thus, by the definition in Eq.~\eqref{eq:app_dom_noise_covariance},
\begin{equation}
C_{s+1}
=
A_cC_sA_c^\top
+
\eta^2\Sigma_{\mathrm{dom}}(\theta_c).
\label{eq:app_covariance_recursion}
\end{equation}
Eq.~\eqref{eq:app_covariance_recursion} is the covariance recursion showing how the dominant noise covariance \(\Sigma_{\mathrm{dom}}(\theta_c)\) induces the local displacement covariance through the local recursion.

\paragraph{Stationary displacement covariance.}
We now derive the stationary displacement covariance created by the local recursion around the reference point \(\theta_c\).

Iterating Eq.~\eqref{eq:app_covariance_recursion} gives
\begin{equation}
C_s
=
A_c^s C_0 (A_c^\top)^s
+
\eta^2
\sum_{\ell=0}^{s-1}
A_c^\ell
\Sigma_{\mathrm{dom}}(\theta_c)
(A_c^\top)^\ell .
\label{eq:app_finite_covariance}
\end{equation}
By Assumption 4 in Appendix~\ref{app:local_assumptions}, \(\rho(A_c)<1\), so
\(A_c^s C_0 (A_c^\top)^s\to 0\), and the second term in Eq.~\eqref{eq:app_finite_covariance} also converges as \(s\to\infty\).
We denote its limit by
\[
C_{\mathrm{mb}}(\theta_c)
:=
\lim_{s\to\infty} C_s
\]
and call it the mini-batch-induced displacement covariance.
Then \(C_{\mathrm{mb}}(\theta_c)\) satisfies
\begin{equation}
C_{\mathrm{mb}}(\theta_c)
=
A_cC_{\mathrm{mb}}(\theta_c)A_c^\top
+
\eta^2\Sigma_{\mathrm{dom}}(\theta_c).
\label{eq:app_cmb_lyapunov}
\end{equation}
Eq.~\eqref{eq:app_cmb_lyapunov} is a discrete Lyapunov equation for \(C_{\mathrm{mb}}(\theta_c)\).
Equivalently, \(C_{\mathrm{mb}}(\theta_c)\) is given by the \(s\to\infty\) limit of Eq.~\eqref{eq:app_finite_covariance}, namely
\begin{equation}
C_{\mathrm{mb}}(\theta_c)
=
\eta^2
\sum_{\ell=0}^{\infty}
A_c^\ell
\Sigma_{\mathrm{dom}}(\theta_c)
(A_c^\top)^\ell .
\label{eq:app_cmb_series}
\end{equation}
Eq.~\eqref{eq:app_cmb_series} shows that \(C_{\mathrm{mb}}(\theta_c)\) is formed by accumulating the dominant noise covariance \(\Sigma_{\mathrm{dom}}(\theta_c)\) through the local recursion, with its value determined by the noise covariance, the local Hessian structure, and the learning rate.
This covariance summarizes the effect of mini-batch noise in the dominant subspace.

\paragraph{Mini-batch-induced sharpness correction.}
Now set \(C=C_{\mathrm{mb}}(\theta_c)\) in Eq.~\eqref{eq:app_avg_gradient_expansion}.
Then the averaged gradient is
\begin{equation}
\mathbb E_\delta[\nabla L(\theta_c+\delta)]
=
\nabla L(\theta_c)
+
\frac12
\left.
\nabla_{\vartheta}
\operatorname{Tr}\!\left(H(\vartheta)C_{\mathrm{mb}}(\theta_c)\right)
\right|_{\vartheta=\theta_c}
+
\mathcal R_3(\theta_c).
\label{eq:app_mb_avg_gradient}
\end{equation}
Eq.~\eqref{eq:app_mb_avg_gradient} shows that the mini-batch-induced displacement covariance \(C_{\mathrm{mb}}(\theta_c)\) leaves an additional term in the averaged gradient.
We refer to this term as the mini-batch-induced sharpness correction term. For the isotropic covariance in Section~\ref{app:dominant_cases}, this term is proportional to
\(\nabla S_k(\theta_c)\).
In contrast, \(C_{\mathrm{mb}}(\theta_c)\) is generally not isotropic, so the correction term in Eq.~\eqref{eq:app_mb_avg_gradient} is not simply proportional to \(\nabla S_k(\theta_c)\).

Therefore, in Section~\ref{sec:experiments}, we apply the corresponding sharpness correction to full-batch GD and compare the resulting corrected GD with mini-batch SGD.
Specifically, Figure~\ref{fig:teaser} compares full-batch GD, mini-batch SGD, and corrected GD, all starting from the same initialization, and shows that this correction reduces the sharpness gap between GD and SGD.

\section{Experimental Details}
\label{app:experimental_setup}

In this section, we summarize the codebase, datasets, architectures,
optimization settings, Hessian computation, and correction computation used in the main experiments.

\subsection{Implementation}
\label{app:implementation}

The experimental code is based on the supplementary implementation provided with
\citet{song2025does}. This implementation follows the experimental setup and codebase of
\citet{cohen2021gradient}. 
The experiments were run on an internal server with \(8\) NVIDIA RTX 3090 GPUs.

On top of this codebase, we added the estimation of the displacement covariance \(C_{\mathrm{mb}}\),
the correction term \(b_{\mathrm{corr}}\), and corrected GD used in this paper.
For Hessian eigenspace computation and higher-order automatic differentiation, we followed
the numerical conventions of \citet{cohen2025understanding}. In particular, for the Transformer experiments, we use a vanilla PyTorch LayerNorm implementation instead of the default PyTorch \texttt{nn.LayerNorm}, which was needed to compute the third-order derivatives in the correction term.

\subsection{Datasets and Architectures}
\label{app:exp_data_models}

Table~\ref{tab:app_datasets_architectures} summarizes the datasets and architectures used in our experiments.
\begin{table}[H]
\centering
\small
\setlength{\tabcolsep}{4pt}
\caption{Dataset--architecture pairs used in our experiments.}
\label{tab:app_datasets_architectures}
\begin{tabular}{@{}p{0.18\textwidth} p{0.12\textwidth} p{0.62\textwidth}@{}}
\toprule
Dataset & Task & Architecture \\
\midrule
MNIST-5k \citep{726791}
&
10-class
&
3-layer MLP with two width-200 hidden layers and \(\tanh\) activation
\\
CIFAR10-5k \citep{krizhevsky2009learning}
&
10-class
&
3-layer CNN with width \(32\), ReLU activation, max pooling, and a linear classifier head
\\
SST2-1k \citep{socher-etal-2013-recursive}
&
Binary
&
2-layer Transformer encoder with hidden dimension \(64\), \(8\) attention heads, ReLU activation, mean pooling, and a linear classifier head
\\
\bottomrule
\end{tabular}
\end{table}

For MNIST-5k and CIFAR10-5k, we use the first \(5{,}000\) training examples.
For SST2-1k, we use the first \(1{,}000\) training examples. All experiments use
only the training split. The MLP and CNN settings follow
\citet{cohen2021gradient}, while the Transformer setting follows
\citet{damian2023selfstabilizationimplicitbiasgradient}.

\subsection{Training Setup}
\label{app:training_protocol}

All main runs use vanilla mini-batch SGD or full-batch GD.
The learning rate is kept constant, and we do not use momentum, weight decay, warmup,
learning-rate decay, or early stopping. The default loss is the MSE loss with one-hot labels.
Gradients, Hessian-vector products, Hessian eigenvalues, and covariance estimates are all
computed using the \(1/N\)-normalized training loss.

Unless stated otherwise, we follow the stable learning-rate (GF) regime of
\citet{song2025does}. We use a batch size of \(50\), with learning rate \(0.01\)
for the MLP and \(0.001\) for the CNN and Transformer. Each main run is trained
for \(20{,}000\) steps.

\subsection{Hessian and Correction Computation}
\label{app:hessian_correction}

Hessian-related quantities are computed at analysis checkpoints placed every \(100\) training steps.
At each checkpoint, we compute the top Hessian eigenpairs using Hessian-vector products and LOBPCG
\citep{doi:10.1137/S1064827500366124}. We compute \(20\) eigenpairs at each checkpoint and use
the top-\(k\) eigenvectors as the dominant-subspace basis. We use \(k=10\) for MNIST-5k and CIFAR10-5k,
and \(k=2\) for SST2-1k.

At the same checkpoint, we estimate the mini-batch noise covariance on the dominant Hessian basis using \(100\) newly sampled mini-batches of size \(50\).
The quantities \(C_{\mathrm{mb}}\) and \(b_{\mathrm{corr}}\) are then computed as described in Appendix~\ref{app:mb_covariance}. In corrected GD,
\(b_{\mathrm{corr}}\) is recomputed at each checkpoint and kept fixed until the next checkpoint.
\subsection{Reported Quantities and Baselines}
\label{app:metrics_baselines}

At the analysis checkpoints above, we record training loss, gradients, top Hessian eigenvalues,
top-\(k\) sharpness \(S_k\), and dominant alignment \(\chi_k\).
The main comparisons are mini-batch SGD, full-batch GD, and corrected GD.
In the projected-update experiments, we also report Dom-SGD and Bulk-SGD.
In the perturbation experiments, we compare dominant perturbation with
random perturbation matched in rank and total variance.

\section{Additional Experiments}
\label{app:additional_experiments}

In this section, we provide additional experimental results complementing
Section~\ref{sec:experiments}. We first examine how the sharpness evolution changes when the batch size is changed.
We then reverse the direction of the correction term \(b_{\mathrm{corr}}\) to check whether the sharpness-reducing effect depends on
the direction of the correction.

\subsection{Batch Size Sweep}
\label{app:batch_size_sweep}
\begin{figure}[H]
  \centering
  \includegraphics[width=0.32\textwidth]{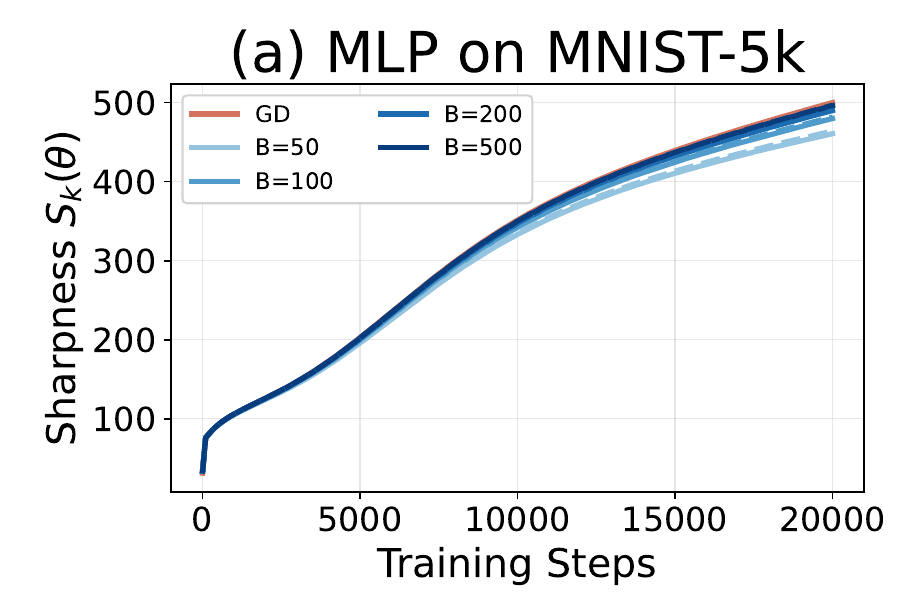}\hfill
  \includegraphics[width=0.32\textwidth]{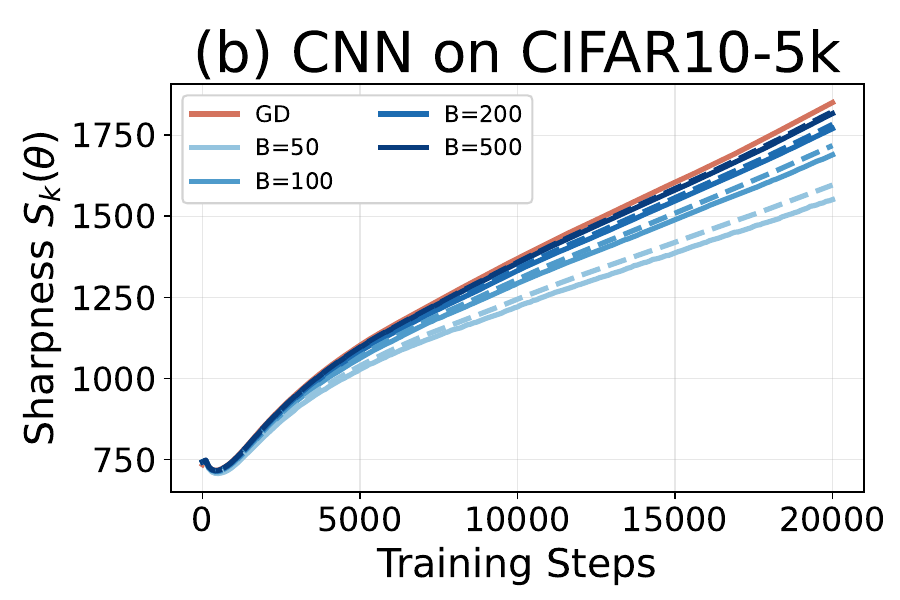}\hfill
  \includegraphics[width=0.32\textwidth]{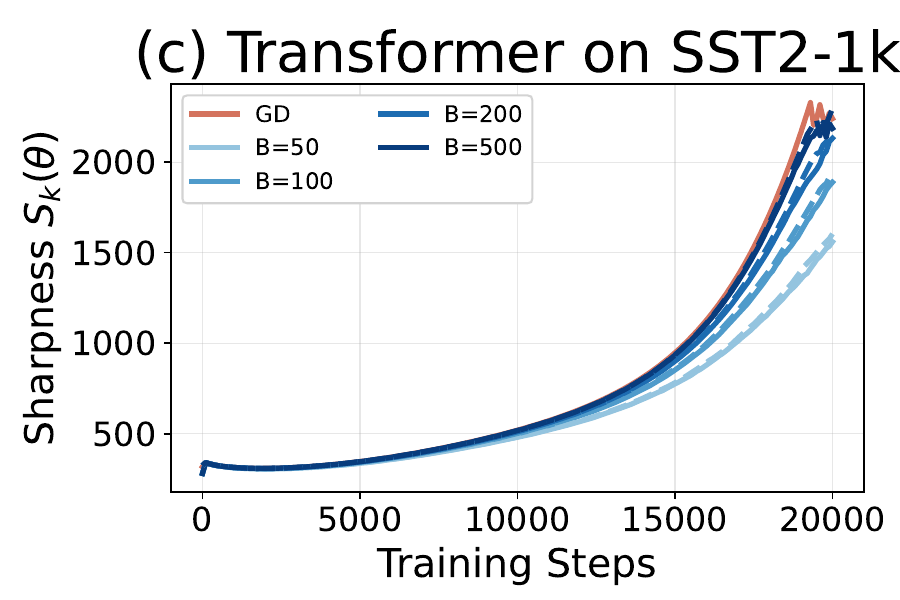}
  \vspace{-1mm}
  \caption{
  \textbf{Corrected GD across batch sizes.}
  (a) MLP on MNIST-5k, (b) CNN on CIFAR10-5k, and (c) Transformer on SST2-1k.
  Solid lines denote mini-batch SGD, and dashed lines denote corrected GD.
  As the batch size increases, the \(S_k\) curve of SGD moves closer to full-batch GD, and corrected GD follows the same trend.
}
\label{fig:batch-size-sweep-plot}
    
\end{figure}

We compare mini-batch SGD, full-batch GD, and corrected GD obtained by applying the sharpness correction term to full-batch GD under different batch sizes in Figure~\ref{fig:batch-size-sweep-plot}.

\subsection{Correction Direction}
\label{app:correction_direction}

\begin{figure}[H]
\centering
    \includegraphics[width=0.82\textwidth]{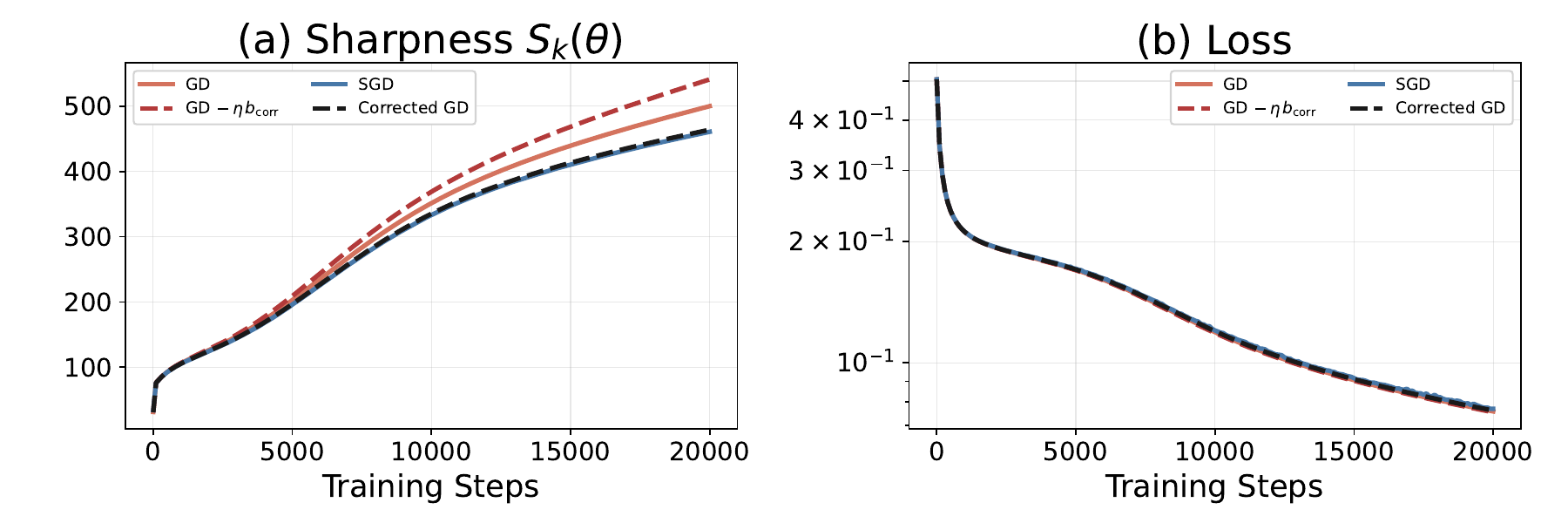}
  \vspace{-1mm}
  \caption{
\textbf{Effect of the correction direction on MLP / MNIST-5k.}
We compare GD, SGD, corrected GD, and GD with the reversed correction.
(a) Top-\(k\) sharpness \(S_k\). (b) Training loss.
The derived correction lowers sharpness relative to GD, whereas reversing its direction increases sharpness relative to GD.
The loss curves remain largely unchanged.
}
  \label{fig:app_correction_direction_mlp}
  \vspace{1.8mm}
  \centering
  \includegraphics[width=0.82\textwidth]{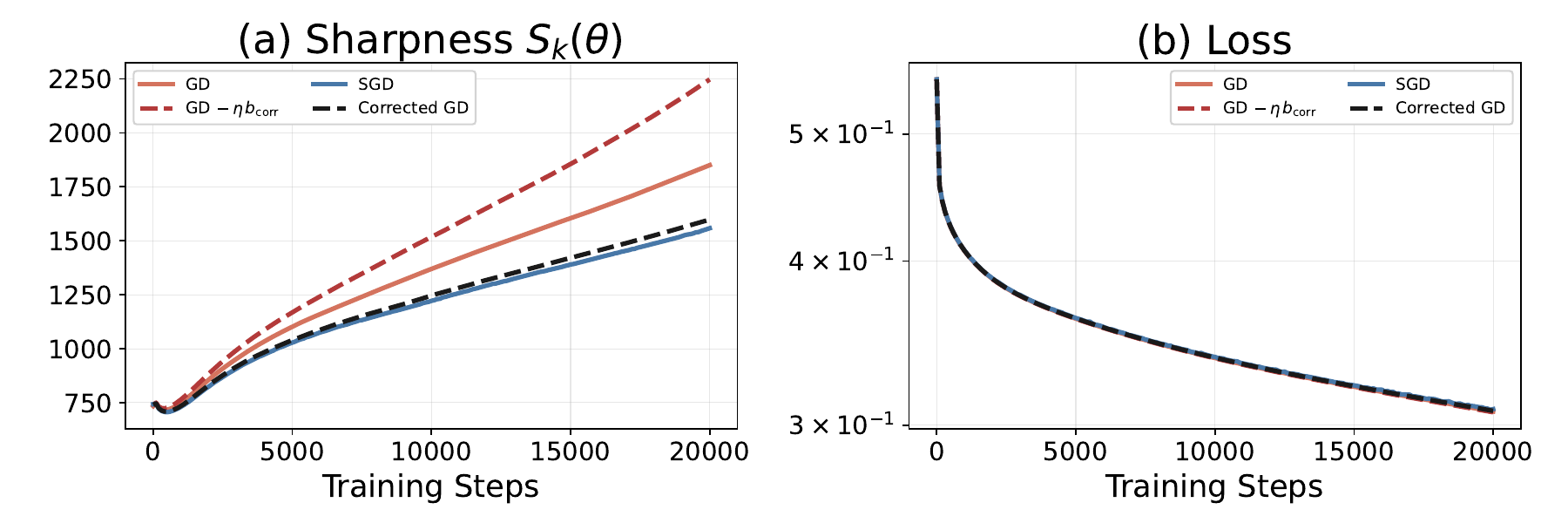}
  \vspace{-1mm}
  \caption{
\textbf{Effect of the correction direction on CNN / CIFAR10-5k.}
We compare GD, SGD, corrected GD, and GD with the reversed correction.
(a) Top-\(k\) sharpness \(S_k\). (b) Training loss.
The derived correction lowers sharpness relative to GD, whereas reversing its direction increases sharpness relative to GD.
The loss curves remain largely unchanged.
}
  \label{fig:app_correction_direction_cnn}
\end{figure}

We reverse the direction of the correction term \(b_{\mathrm{corr}}\) and compare it with the original corrected GD.
This experiment shows that sharpness reduction depends on the direction of the derived correction term.
Figures~\ref{fig:app_correction_direction_mlp} and~\ref{fig:app_correction_direction_cnn}
show the corresponding sharpness and loss curves.
\vspace{-1mm}
\vspace*{-.5cm}
\end{document}